\documentclass{article}

\usepackage[final,nonatbib]{neurips_2020}

\usepackage{amssymb}
\usepackage[utf8]{inputenc} 
\usepackage[T1]{fontenc}    
\usepackage{hyperref}       
\usepackage{url}            
\usepackage{booktabs}       
\usepackage{amsfonts}       
\usepackage{nicefrac}       
\usepackage{microtype}      
\usepackage{mathrsfs}
\usepackage{amsthm}
\usepackage{amssymb}
\usepackage[tbtags]{amsmath}
\usepackage{mathrsfs}
\usepackage{graphicx}
\usepackage{algorithm}
\usepackage{algorithmic}
\usepackage{subfigure}
\usepackage{listings}
\usepackage{booktabs}
\usepackage{appendix}
\usepackage{multirow}
\usepackage{color}
\usepackage{bm}

\usepackage{titlesec}
\usepackage{titletoc}
\usepackage{xcolor}
\usepackage{wrapfig}

\titlecontents{section}
[1.5cm]
{\bf }%
{\contentslabel{2.5em}}%
{}%
{\titlerule*[0.5pc]{$\cdot$}\contentspage\hspace*{2cm}}%

\title{Select-ProtoNet: Learning to Select \\for Few-Shot Disease Subtype Prediction}

\author{
  Ziyi Yang\\
  Macau University of Science and Technology\\
  \texttt{yangziyi091100@163.com}\\
  \And
  Jun Shu\\
  Xi’an Jiaotong University\\
  \texttt{xjtushujun@gmail.com}\\
  \And
  Yong Liang\\
  Macau University of Science and Technology\\
  \texttt{yliang@must.edu.mo}\\
  \And
  Deyu Meng\\
  Xi’an Jiaotong University\\
  \texttt{dymeng@mail.xjtu.edu.cn}\\
  \And
  Zongben Xu\\
  Xi’an Jiaotong University\\
  \texttt{zbxu@mail.xjtu.edu.cn}\\
}

\begin{document}

\maketitle

\begin{abstract}
Current machine learning has made great progress on computer vision and many other fields attributed to the large amount of high-quality training samples, while it does not work very well on genomic data analysis, since they are notoriously known as small data. In our work, we focus on few-shot disease subtype prediction problem, identifying subgroups of similar patients that can guide treatment decisions for a specific individual through training on small data. In fact, doctors and clinicians always address this problem by studying several interrelated clinical variables simultaneously. We attempt to simulate such clinical perspective, and introduce meta learning techniques to develop a new model, which can extract the common experience or knowledge from interrelated clinical tasks and transfer it to help address new tasks. Our new model is built upon a carefully designed meta-learner, called Prototypical Network \cite{snell2017prototypical}, that is a simple yet effective meta learning machine for few-shot image classification. Observing that gene expression data have specifically high dimensionality and high noise properties compared with image data, we proposed a new extension of it by appending two modules to address these issues. Concretely, we append a feature selection layer to automatically filter out the disease-irrelated genes and incorporate a sample reweighting strategy to adaptively remove noisy data, and meanwhile the extended model is capable of learning from a limited number of training examples and generalize well. Simulations and real gene expression data experiments substantiate the superiority of the proposed method for predicting the subtypes of disease and identifying potential disease-related genes.
\end{abstract}

\section{Introduction}
\label{Introduction}
Disease subtype prediction is to identify subgroups of similar patients that can guide treatment decisions for a specific individual \cite{saria2015subtyping, wang2014similarity}. For instance, in the past 15 years, five subtypes of breast cancer have been identified and intensively studied \cite{sohn2017clinical}. At the level of molecular biology, the use of gene expression data to predict disease subtypes is of great significance for improving the accuracy of disease diagnosis and identifying potential disease-related genes. However, one of the challenging problem is that the gene expression data are notoriously known as small data \cite{shu2019meta}, i.e., we only have a relatively small number of samples for each disease subtype. The small data learning or few-shot learning has recently attracted many researches in machine learning community \cite{shu2019meta,snell2017prototypical}. This inspires us to bring the state-of-the-art algorithms of few-shot learning to genomic data analysis.

In our paper, we attempt to deal with few-shot disease subtype prediction problem. The most used method in genomic data is to augment sample size by aggregating data from multiple studies under comparable conditions or treatments \cite{li2014meta,hughey2015robust,yang2019multi}. This strategy commonly encounters bottlenecks due to the complex properties of gene expression data, in which the aggregation of data from different platforms or experiments inevitably suffers from batch effects, heterogeneity, and other sources of bias \cite{lazar2012survey}. Besides, we observe that the current methods usually consider only the subtype prediction task for a particular disease, without taking into account several clinical variables simultaneously that are often of concern to both doctors and clinicians. To simulate the process of doctors and clinicians studying disease subtype prediction, we introduce the meta learning techniques to develop a new data efficient model, which can extract the shared experience or knowledge from a series of related tasks, and rapidly transfer it to the new tasks \cite{finn2017model,shu2018small,hospedales2020meta}. Therefore, the basic idea of the proposed new model is to learn from interrelated clinical tasks through meta learning techniques to extract the valuable information to help model generalize to the disease subtype prediction task well.

Prototypical Network (ProtoNet) \cite{snell2017prototypical} is a carefully designed meta-learner that is a simple and yet effective meta learning machine for few-shot image classification. It tries to learn a metric space in which classification can be performed by computing distances to prototype representations of each class. However, unlike images, gene expression data are much harder to analyze due to their high-dimensional and high-noise properties. The curse of dimensionality problems tends to make predictions become more challenging since a large number of redundant features involving in decision. Besides, existing evidence suggests that a high level of technical or biological noise inevitably exists in gene expression data \cite{raser2005noise,schmiedel2019empirical}, which tends to easily occur overfitting issue and lead to poor generalization performance. To address these issues, we propose Select-ProtoNet, a new extension of ProtoNet by appending two modules, feature selection layer and sample selection net, making it filter out the disease-irrelated genes and remove noisy data towards few-shot disease subtype prediction.

Our proposed method is built on top of ProtoNet,
and we additionally append a feature selection layer to automatically cherry-pick the disease-related genes, and incorporate a sample reweighting strategy to adaptively suppress the negative influence of noisy data. With these two modules, our method can perform robust to high-dimensional and high-noise gene expression data, and allows for better generalization on small data benefited from ProtoNet.

\begin{figure}[!pbt]
	\centering
	\includegraphics[width=5in]{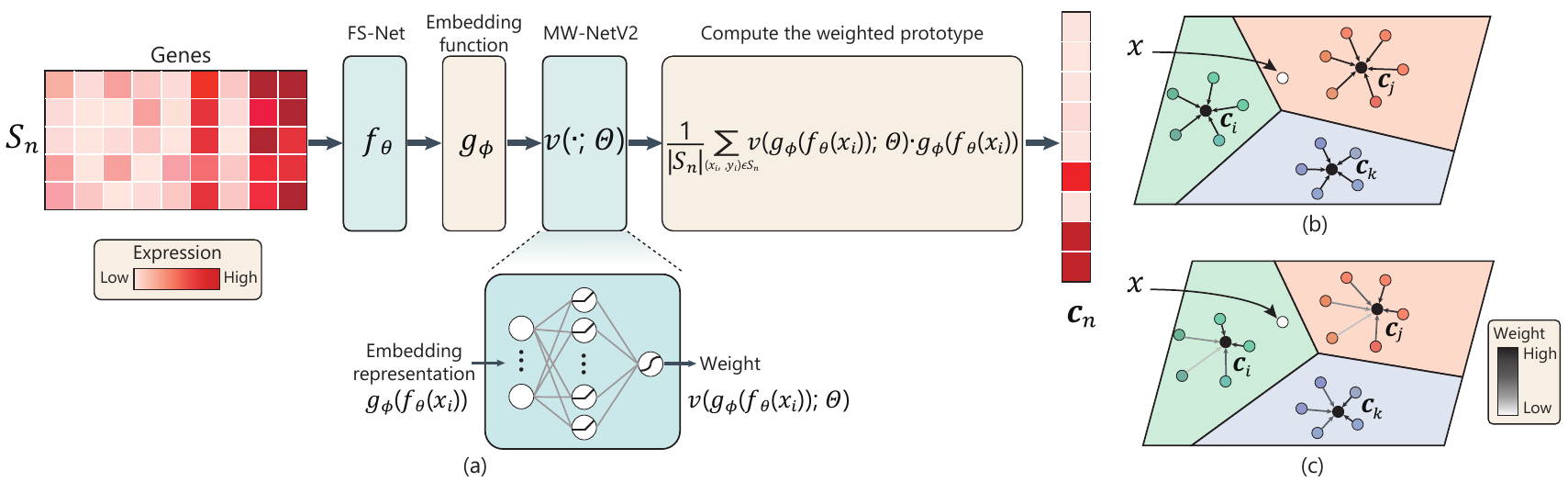} \vspace{-2mm}
	\caption{\emph{(Left)} (a) Diagram of the Select-ProtoNet structure. The weighted prototype $\mathbf{c}_{n}$ is calculated by weighted representation of all the embedded support samples for class $n$. The blue boxes represent the newly appending modules to ProtoNet. \emph{(Right)} Qualitative example of how Select-ProtoNet works. Assume a query sample $x$ belongs to class $i$. (b) The prototypes $\mathbf{c}_{n}$ are calculated by the mean of embedded support samples for each class. The closet prototype to the query sample $x$ is $\mathbf{c}_{j}$. (c) The prototypes $\mathbf{c}_{n}$ are calculated by weighted representation of all the embedded support samples for each class. The closet prototype to the query sample is now the one of the class $i$.}
	\label{Fig:00}  \vspace{-4mm}
\end{figure}

The key contributions of this paper can be summarized as follows:
 \begin{itemize}
	\item We propose a new extensions of ProtoNet, called Select-ProtoNet, for few-shot disease subtype prediction, as shown in Fig.~\ref{Fig:00}(a). To our best knowledge, this should be the first work to use meta-learning techniques for Bioinformatics. By additionally appending two modules, feature selection layer and sample selection net, our model can adaptively select important features and clean samples, as well as learn from small data and generalize well.
	\item The additionally appending modules and the vanilla ProtoNet are updated in a unified framework, which can be easily implemented on the basis of the ProtoNet.
	\item
	To address the absence of benchmark datasets for testing few-shot classification in the field of genomics, we devise a new dataset, \emph{mini}TCGA Meta-Dataset, which is derived from TCGA Meta-Dataset \cite{samiei2019tcga}. This dataset can be used as a benchmark for meta learning methods in genomics applications under few-shot learning setting.
	\item We experimentally show the superiority of the proposed method for predicting the subtypes of disease and identifying potential disease-related genes.
\end{itemize}

The remaining sections are organized as follows. Section \ref{Related_Work} reviews the related work. Section \ref{Method} detailedly introduces the proposed method. Section \ref{Experiments} demonstrates simulation and real data experimental results and the conclusion is finally made.

\section{Related Work}
\label{Related_Work}
\textbf{Data Integration.}
Genomics data integration is usually used method to deal with small sample problem \cite{lazar2013batch}. It can be roughly performed through two different strategies: (1) `meta-analysis' \cite{rhodes2004large, rhodes2005integrative, wirapati2008meta} analyzes each dataset independently and finally combines the statistic results to identify disease-related genes; (2) `integrative analysis via data merging' \cite{shabalin2008merging,leek2010tackling,hughey2015robust} first aggregates samples from different datasets into a unified large dataset and then analyzes the new integrated dataset. However, integrating data from different experiments or platforms for integrative analysis easily suffers from batch effects and heterogeneity, which is still very challenging in computational biology.

\textbf{Few-shot Learning.}
Meta-learning has a prominent history in machine learning \cite{bengio1992optimization, schmidhuber1987evolutionary,thrun1998lifelong}, and many meta learning methods have been applied to address the small data learning or few-shot learning \cite{shu2018small}. They can be roughly divided into two main types: (1) gradient-based optimization methods \cite{finn2017model,finn2018probabilistic, ravi2016optimization, lee2018gradient} learn a meta-learner in the outer loop to initialize a base-learner for the inner loop that is then trained on a novel few-shot task; (2) metric-based methods \cite{vinyals2016matching, snell2017prototypical, ren2018meta, wang2018low, xing2019adaptive} learn a metric space of sample features in which classification can be efficient with few samples. Most current methods rely solely on image classification tasks. There is a lack of systematic researches about using meta-learning for few-shot disease subtype prediction problems. In this work, we extend a well-known meta-learner, Prototypical Networks \cite{snell2017prototypical}, to address biological problems. Our model appends two modules that performs robust to high-dimensional and high-noise gene expression data.

\textbf{Learning with High-Dimensional Features.}
Feature extraction and feature selection are two commonly used manners to address the curse of dimensionality problems. Feature extraction transforms the original features into a set of new features through subspace learning \cite{mandanas2020subspace}. However, a reasonable biological interpretation is difficult to obtain from the learned feature subspace. Feature selection removes irrelevant, redundant data, and selects a set of important features that are significantly related to the objective \cite{lazar2012survey}. Regularization technique is a typical feature selection method \cite{tibshirani1996regression, liang2013sparse, xu20101, zou2005regularization}. However, they usually have assumptions about the prior distribution of the data \cite{xue2015survey} and inevitably involve hyper-parameters (i.e., regularization parameters) required to be tuned by cross-validation.

\textbf{Sample Reweighting for Noisy Samples.}
The main idea of sample reweighting strategy is to impose weights on samples based on their reliability for training.
Typically methods include self-paced learning (SPL) \cite{kumar2010self} and its variations \cite{jiang2014easy, yang2019mspl}, iterative reweighting \cite{de2003framework, zhang2018generalized}, FWL \cite{dehghani2017fidelity}, L2RW \cite{ren2018learning}, and Meta-Weight-Net (MW-Net) \cite{shu2019meta}. They incline to suppress the effects of samples with extremely large loss values, possibly with high noisy samples.

\section{Proposed Select-ProtoNet for Few-Shot Disease Subtype Prediction.}
\label{Method}
In this section, we first present preliminaries in the meta learning setting with independent episodic tasks and the prototypical networks (ProtoNet) method for few-shot learning, then give an overview  of our approach, finally elaborate the details of our method.
\subsection{Meta-Learning for Few-Shot Learning}
Few-shot learning (FSL) involves building a model using available training data of seen classes that can classify unseen novel classes using only few examples. Under the FSL setting, we have a large sample set $\mathcal{D}_s$ from a set of source classes $\mathcal{C}_s$, while a few-shot sample set $\mathcal{D}_t$ from a set of target classes $\mathcal{C}_t$, and a test set $\mathcal{T}$ from $\mathcal{C}_t$, where $\mathcal{C}_s \cap \mathcal{C}_t = \emptyset$. The goal of the FSL is to training a classification model with $\mathcal{C}_s$ that can generalize well to $\mathcal{T}$.

We then define the episodic training strategy widely used by existing meta-learning based FSL models \cite{vinyals2016matching,snell2017prototypical}. Concretely, we can define a series of $n$-way $k$-shot tasks randomly sampled from $\mathcal{D}_s$, and each $n$-way $k$-shot task is defined as an episode $D = (S,Q)$. Generally, we call $S$ the support set containing $n$ classes and $k$ samples per class, and $Q$ the query set with the same $n$ classes. In our paper, we try to construct a new meta dataset of FSL for disease subtype prediction in biological field. The episode $D$ can be constructed by the following process: we first select a small set of source class containing $n$ classes $\mathcal{C}=\{\mathcal{C}_i|i=1,2,\cdots,n_{c}\}$ from $\mathcal{C}_s$, and then generate $S$ and $Q$ by randomly sampling $k$ support samples and $q$ query samples from each class in $C$, respectively. Therefore, we have $S=\{(x_i,y_i)|y_i\in C,i=1,2,\cdots,m_s\}$, and $Q=\{(x_i,y_i)|y_i\in C,i=1,2,\cdots,m_q\}$, where $m_s = n\times k$, $m_q$ is the number of samples for query set in each episodic, and $S\cap Q=\emptyset$.

We denote the classifier we want to learn as $p_{\phi} (y|x)$, with parameter $\phi$, which outputs a probability of a data point belonging to the class $y$ given the data sample $x$. The classifier is trained between its predicted labels and the ground truth labels over the query set $Q$ with different episodes such that it can generalize to other datasets, the objective function is written as follows:
\begin{equation}
\phi = \arg\min_{\phi} \mathbb{E}_{\mathcal{D}\subset \mathcal{D}_s} \sum_{(x,y)\in Q} -\log p_{\phi}(y|x,S),
\end{equation}
where different forms of $p_{\phi}(y|x,S)$ define different kinds of meta learning algorithm.

\subsection{Review of Prototypical Networks}
Here, we will give a introduction to the prototypical networks (ProtoNet), which is a well-known meta-learning algorithm, and our model is built on top of it. ProtoNet learns a prototype of each class in the support set $S$ and classifies each sample in the query set $Q$ based on its distances to different prototypes.
Thus $p_{\phi}(y|x,S)$ in ProtoNet is defined as follows:
\begin{align}\label{Eq1}
p_{\phi}(y| x, S)=\frac{\exp \left(-d\left(g_{\phi}(x), \mathbf{c}_{n}\right)\right)}{\sum_{n^{\prime}} \exp \left(-d\left(g_{\phi}(x), \mathbf{c}_{n^{\prime}}\right)\right)},
\end{align}
where $d$ represents the Euclidean distance metric in the feature space, and $\mathbf{c}_{n}$ is prototype for every episode. Eq.\ref{Eq1} means that ProtoNet produces the class distributions of a query sample $x$ based on the softmax output w.r.t. the distance between $g_{\phi}(x)$ and the class prototype $\mathbf{c}_{n}$, where $g_{\phi}(x)$ is the embedding function which maps a sample $x$ to the feature vector $g_{\phi}(x)$.
Each prototype $\mathbf{c}_{n}$ is calculated by averaging the vectors of all the embedded support samples $\mathcal{S}_{n}$ belonging to the class $n$, given by the following form:
\begin{equation}\label{Eq_02}
\mathbf{c}_{n}=\frac{1}{\left|\mathcal{S}_{n}\right|} \sum_{\left(x_{i}, y_{i}\right) \in \mathcal{S}_{n}} g_{\phi}\left(x_{i}\right),
\end{equation}
where $\mathcal{S}_{n}\in S$ denotes the set of support samples that belong to the class $n$.

\subsection{Learning to Select Important Features and Clean Data for Disease Subtype Prediction}
We attempt to introduce ProtoNet to disease subtype prediction problem to enhance the capability of few-shot recognition. However, the gene expression data is harder to deal with compared with image dataset, since for its high dimensionality and high noise properties. To further overcome these issues, we proposed to append two modules (as shown in Fig.\ref{Fig:00} (a)), i.e.,
a feature selection layer (FS-Layer) to automatically filtering out the disease-irrelated genes and incorporate an adaptive sample reweighting strategy (MW-NetV2) against data noise, building up on the ProtoNet.

\subsubsection{Feature Selection Layer}
For each sample $x \in \mathbb{R}^p$ of gene expression data, the feature dimension $p$ is high. Generally, feature selection methods try to find a selection vector $\beta=(\beta_1,\beta_2,\cdots,\beta_p)$, which is element-wise multiplication with data $x$, to filter out the useless feature and obtain a new representation of data $x_{new}$ to help the following tasks perform well, i.e.,
\begin{align}
x_{new} = \beta \odot x, \beta_j \in [0,1].
\end{align}
The regularization technique, the effective method on this problem, often manually set a specific form of regularization under a certain assumption on training data, which is infeasible when we know little knowledge underlying gene expression data. To overcome this, we model the selection vector as a Softmax network layer as follows, which can learn an adaptive feature weighting vector from data,
\begin{align}
x_{new} = \beta(\theta) \odot x \triangleq f_{\theta}(x), \beta_i(\theta)=\frac{\exp(\theta)_i}{\sum_{j}\exp(\theta)_j}, \sum_{i}\beta_i(\theta)=1,
\end{align}
where $\theta \in \mathbb{R}^p$ is the parameter vector of Softmax layer, and $\exp$ is the element-wise exponential operator.
It can be be easily embedded into Eq.(\ref{Eq1}),
\begin{align}\label{Eq2}
p_{\phi,\theta}(y| x, S)=\frac{\exp \left(-d\left(g_{\phi}(f_{\theta}(x)), \mathbf{c}_{n}\right)\right)}{\sum_{n^{\prime}} \exp \left(-d\left(g_{\phi}(f_{\theta}(x)), \mathbf{c}_{n^{\prime}}\right)\right)},
\end{align}
where $\mathbf{c}_{n}$ can be rewritten as:
\begin{equation}\label{Eq_03}
\mathbf{c}_{n}=\frac{1}{\left|\mathcal{S}_{n}\right|} \sum_{\left(x_{i}, y_{i}\right) \in \mathcal{S}_{n}} g_{\phi}\left(f_{\theta}(x_i)\right),
\end{equation}
This formulation needs no expert-level knowledge to character $\beta$, and $\theta$ can efficiently learned by automatic differentiation techniques as $\phi$ does,  which makes it easy to scale to other problems.

\begin{algorithm}[!ht] 
	\caption{Training episode loss computation for Select-ProtoNet. $N$ is the number of examples in the training set, $K$ is the number of classes in the training set, $n_{c}\leq K$ is the number of classes per episode, $k$ is the number of support examples per class, $q$ is the number of query	examples per class. \textsc{RandomSample}($S$,$n$) denotes a set of $n$ elements chosen uniformly at random from set $S$, without replacement. \textcolor{blue}{\emph{This algorithm is built upon ProtoNet \cite{snell2017prototypical}, and the revised parts are shown in blue.}}}
	\label{Alg:01}
	\begin{algorithmic}[1]
		\REQUIRE Training set $\mathcal{D} = \{(x_1, y_1), \ldots, (x_N, y_N) \}$, where each $y_i \in \{1, \ldots, K\}$. $\mathcal{D}_k$ denotes the subset of $\mathcal{D}$ containing all elements $(x_i, y_i)$ such that $y_i = n$.
		\ENSURE The loss $\mathcal{J}(\phi,\theta,\Theta)$ for a randomly generated training episode. \\
		\STATE $V \gets \textsc{RandomSample}(\{1,\dots,K\},n_{c})$  \COMMENT{Select class indices for episode}
		\FOR{$n$ in \{$1, \dots, n_{c}$\}}
		\STATE $\mathcal{S}_{n} \gets \textsc{RandomSample}(\mathcal{D}_{V_{n}},k)$  \COMMENT{Select support examples}
		\STATE $\mathcal{Q}_{n} \gets \textsc{RandomSample}(\mathcal{D}_{V_{n}} \setminus \mathcal{S}_{n}, q)$  \COMMENT{Select query examples}
		\STATE \textcolor{blue}{$\mathbf{c}_{n} \gets \frac{1}{\left|\mathcal{S}_{n}\right|} \sum_{\left(x_{i}, y_{i}\right) \in \mathcal{S}_{n}} \mathcal{V}(g_{\phi}(f_{\theta}(x_i));\Theta) \cdot g_{\phi}\left(f_{\theta}(x_i)\right)$ \COMMENT{Compute the weighted prototype}}
		\ENDFOR \\
		\STATE $\mathcal{J}(\phi,\theta,\Theta) \gets 0$  \COMMENT{Initialize loss}
		\FOR{$n$ in \{$1, \dots, n_{c}$\}}
		\FOR{$(x,y)$ in $\mathcal{Q}_{n}$}
		\STATE \textcolor{blue}{$\mathcal{J}(\phi, \theta, \Theta) \gets \mathcal{J}(\phi, \theta, \Theta)+\frac{1}{n_{c} \cdot q}\left[d\left(g_{\phi}\left(f_{\theta}(x)\right), \mathbf{c}_{n}\right)+\log \sum_{n^{\prime}} \exp \left(-d\left(g_{\phi}\left(f_{\theta}(x)\right), \mathbf{c}_{n^{\prime}}\right)\right)\right]$}  \\ \textcolor{blue}{\COMMENT{Update loss}}
		\ENDFOR
		\ENDFOR
	\end{algorithmic}
\end{algorithm}\vspace{-4mm}
\subsubsection{Sample Selection Net}
There exist high noise in gene expression data, which can easily lead to poor performance in generalization. To overcome this, instead of the simple average strategy in ProtoNet, we attempt to assign weights to support samples to character the clean confidence of data, expecting to suppress the effects of samples with extremely noise. Specifically, the final prototype $\mathbf{c}_{n}$ should be calculated by weighted representation of all the
embedded support samples for class $n$, i.e.,
\begin{align}\label{eq5}
\mathbf{c}_{n}=\frac{1}{\left|\mathcal{S}_{n}\right|} \sum_{\left(x_{i}, y_{i}\right) \in \mathcal{S}_{n}} v_i \cdot g_{\phi}\left(f_{\theta}(x_i)\right), v_i\in[0,1],
\end{align}
where $v_i$ reflect the confidence of the support sample $x_i$ belonging to clean data. In general, large weights $v$ are more likely to be high-confident ones with clean data.

To determine the $v$, inspired by current adaptive sample weighting strategy Meta-Weight-Net \cite{shu2019meta}, we attempt to learn a weighting function to distinguish clean and noisy data. Specifically, MW-Net models the sample weights $v$ as an MLP network $\mathcal{V}(\ell;\Theta)$ with only one hidden layer, which is a universal approximator for almost any continuous function and thus can fit a wide range of weighting functions. Its input is the loss function of the sample, and output is the weight to this sample. Since ProtoNet does not compute the losses of the support samples, we use the embedding representation $g_{\phi}\left(f_{\theta}(x)\right)$ as input of MW-Net. Therefore, the Eq.(\ref{eq5}) function can be furtherly rewritten as:
\begin{align}\label{eq6}
\mathbf{c}_{n}=\frac{1}{\left|\mathcal{S}_{n}\right|} \sum_{\left(x_{i}, y_{i}\right) \in \mathcal{S}_{n}} \mathcal{V}(g_{\phi}(f_{\theta}(x_i));\Theta) \cdot g_{\phi}\left(f_{\theta}(x_i)\right),
\end{align}
where $\Theta$ represents the parameters of MW-NetV2 (to distinction with the original MW-Net), and the architecture of MW-NetV2 is same with MW-Net, except for the input. The input of MW-NetV2 is the embedding representation for each sample instead of the loss in MW-Net.

\subsubsection{Learning Algorithm}
By additionally appending feature selection layer (FS-Net) and sample selection net (MW-NetV2) to vanilla ProtoNet, the proposed method, Select-ProtoNet, can simultaneously cherry-pick the disease-related genes helping classification and suppress the negative influence of noisy data. The learning procedure is similar to vanilla ProtoNet, and the parameters of FS-Net and MW-NetV2 are simultaneously updated with the parameters of ProtoNet. The final loss function is:
\begin{align}
\phi,\theta,\Theta =  \arg\min_{\phi,\theta,\Theta} \mathbb{E}_{\mathcal{D}\subset \mathcal{D}_s} &\sum_{(x,y)\in Q} -\log p_{\phi,\theta,\Theta}(y|x,S), \\
p_{\phi,\theta,\Theta}(y|x,S)=\frac{\exp \left(-d\left(g_{\phi}(f_{\theta}(x)), \mathbf{c}_{n}\right)\right)}{\sum_{n^{\prime}} \exp \left(-d\left(g_{\phi}(f_{\theta}(x)),  \mathbf{c}_{n^{\prime}}\right)\right)},&\mathbf{c}_{n}=\frac{1}{\left|\mathcal{S}_{n}\right|} \sum_{\left(x_{i}, y_{i}\right) \in \mathcal{S}_{n}} \mathcal{V}(g_{\phi}(f_{\theta}(x_i));\Theta) \cdot g_{\phi}\left(f_{\theta}(x_i)\right).
\end{align}
Fig.~\ref{Fig:00}(right) illustrates an example on how the proposed method works. The pseudocode for calculating the episode training loss $\mathcal{J}(\phi,\theta,\Theta)$ is provided in Algorithm 1.

\section{Experiments}\vspace{-2mm}
\label{Experiments}
To evaluate the capability of the proposed method, Select-ProtoNet, we conduct experiments on simulated data sets and real gene expression datasets. We show that Select-ProtoNet outperforms its backbone methods and several conventional methods.

\subsection{Simulation Experiments}
\label{Simulation_Experiments}

\textbf{Simulated Datasets.}
We construct two sets of data $\mathcal{D}_{train}$ and $\mathcal{D}_{test}$ with distinct sets of classes under the FSL setting. The generation of simulated data is refer to the work in \cite{ma2019affinitynet}. For more detailed information on generating simulated datasets, see Appendix A.

\textbf{Noise and Feature Dimension Settings.}
We consider four-level settings of corrupted class labels on the support set. The class label of each support sample is independently changed to a random class with probability $p$, where $p=0\%,10\%,30\%,50\%$. Moreover, we consider four-level settings of irrelevant feature with $d$ dimension, where $d= 100, 500, 1000, 2000$.

\textbf{Baselines.}
We compare against two family of methods. The first is metric-based meta learning methods: ProtoNet and ProtoNet with either of the two appended modules. `SelectF-ProtoNet' and `SelectS-ProtoNet' are models with feature selection layer and sample selection net, respectively.
The second is the conventional methods, usually used to analyze gene expression data to predict disease subtypes. The conventional comparison methods include: Support Vector Machines (SVM), Naive Bayes, Logistic Regression, Logistic regression with Lasso penalty (Logistic\_lasso), Random Forest, NeuralNet \cite{ma2019affinitynet}, and AffinityNet \cite{ma2019affinitynet}. Details of baseline implementations, training and test procedures can be found in Appendix B.

\begin{figure}[!pbt]
\begin{minipage}[t]{0.3\linewidth}
\centering
\includegraphics[width=1.5in]{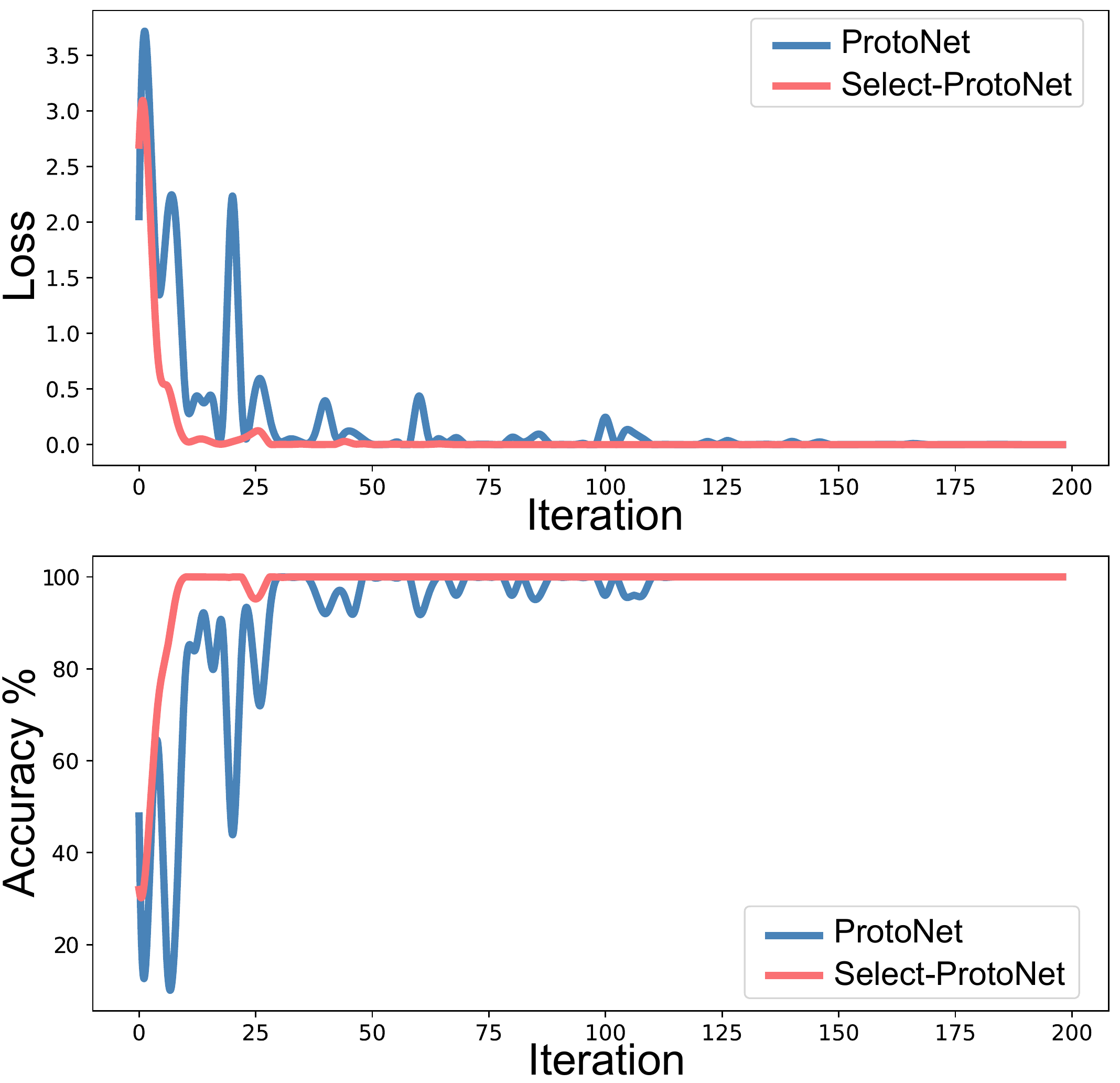}
\caption{Training loss and accuracy (\%) vs. number of iterations under unbiased data.}
\label{Fig:03}
\end{minipage}
\hspace{1mm}
\begin{minipage}[t]{0.3\linewidth}
\centering
\includegraphics[width=1.5in]{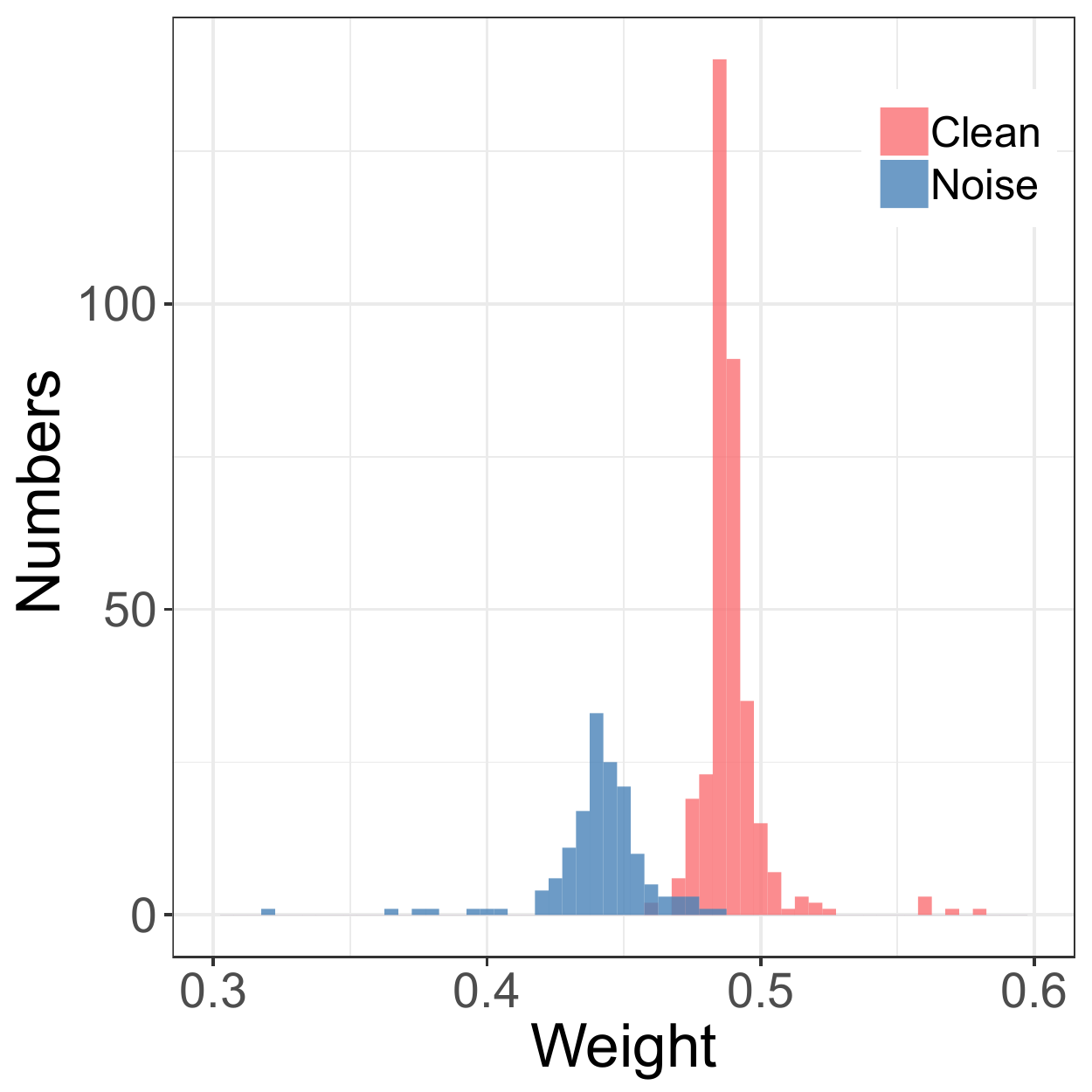}
\caption{Sample weight distribution on training data under 30\% noise rate.}
\label{Fig:07}
\end{minipage}
\hspace{1mm}
\begin{minipage}[t]{0.35\linewidth}
\centering
\includegraphics[width=1.5in]{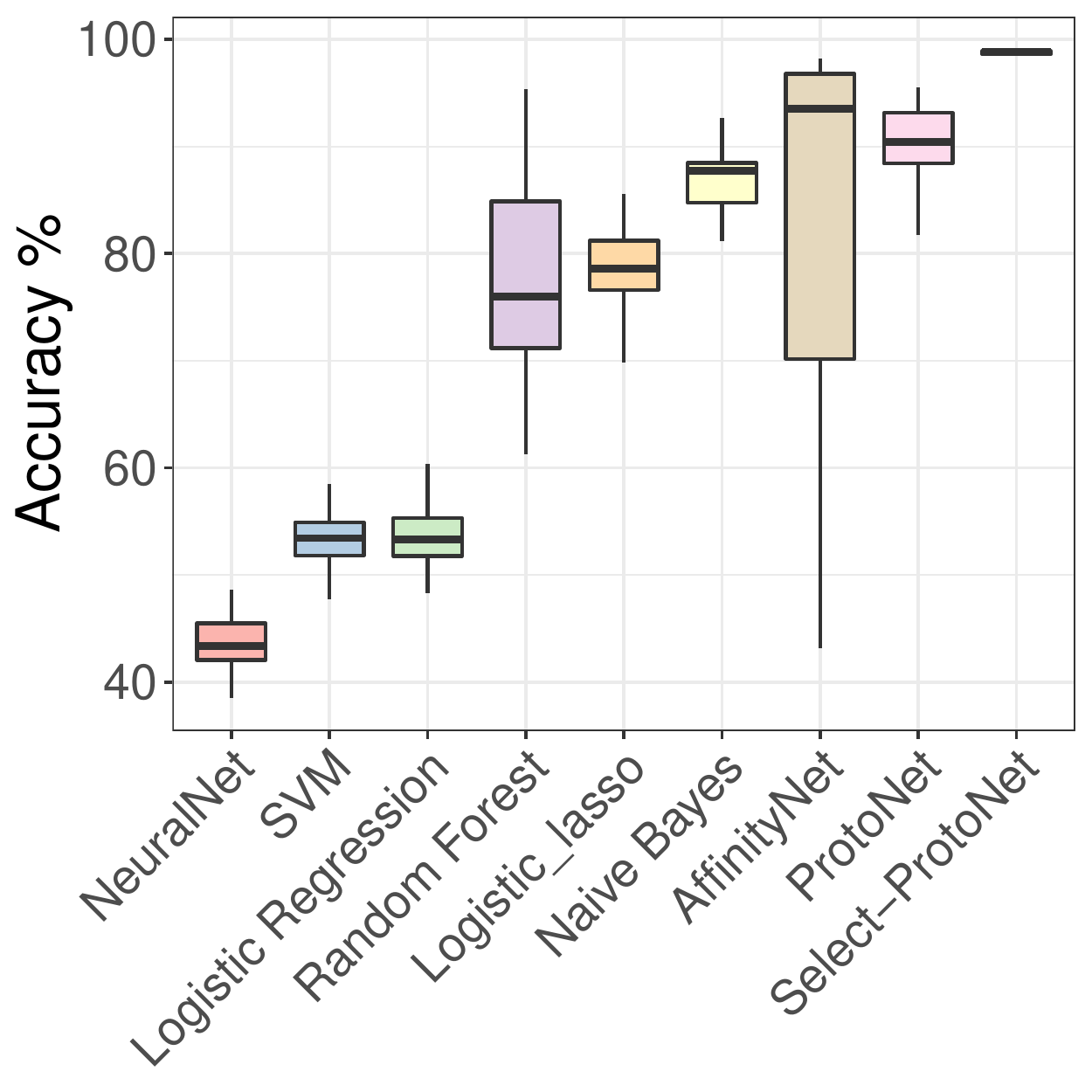}
\caption{Box-plot diagram of test accuracy (\%) of all competing methods over $30$ repetitions.}
\label{Fig:04}
\end{minipage}
\end{figure}

\begin{figure}[!pbt]
\centering
\subfigure[NeuralNet]{
\begin{minipage}[t]{0.28\linewidth}
\centering
\includegraphics[width=1.4in]{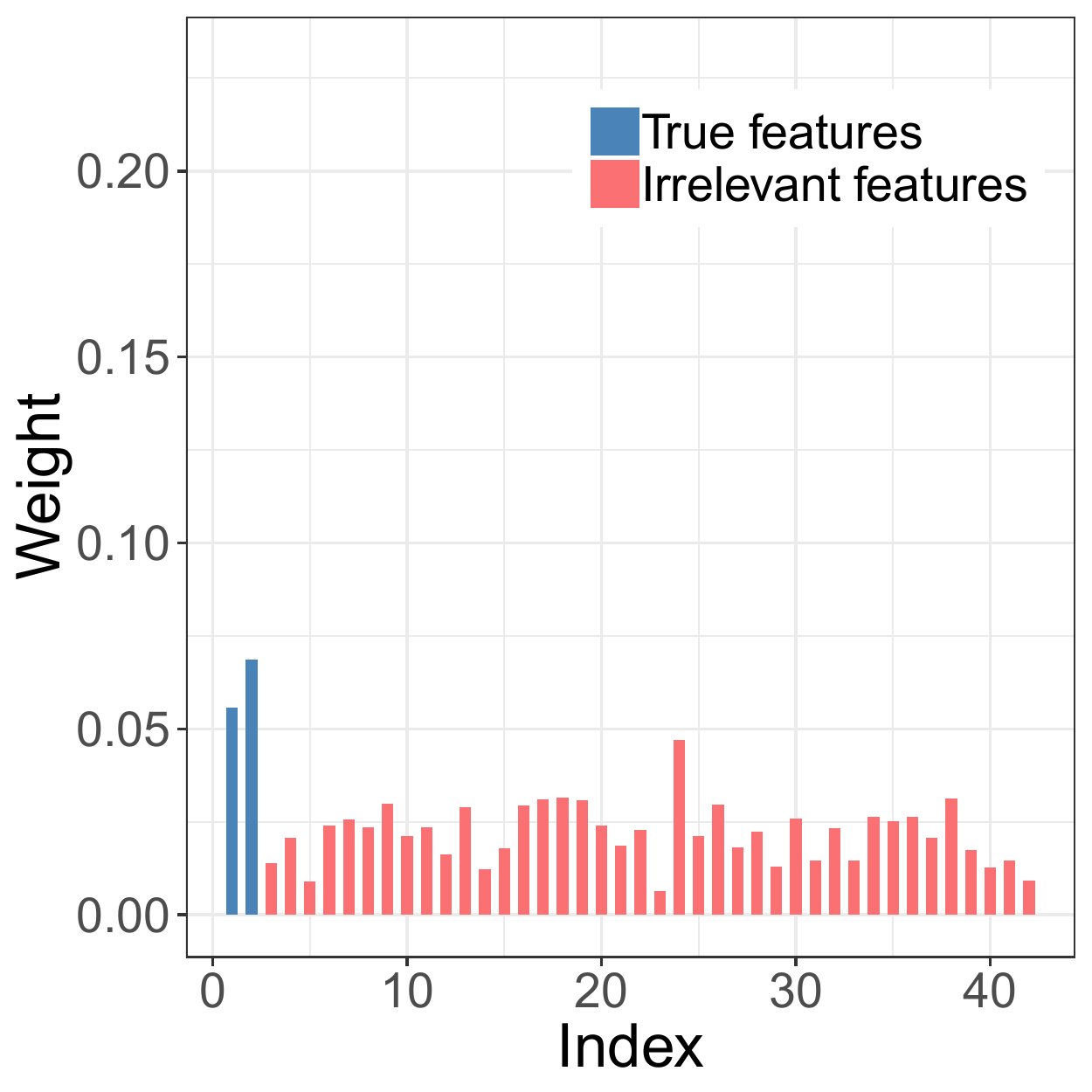}
\label{Fig:05a}
\end{minipage}}
\hspace{4mm}
\subfigure[AffinityNet]{
\begin{minipage}[t]{0.28\linewidth}
\centering
\includegraphics[width=1.4in]{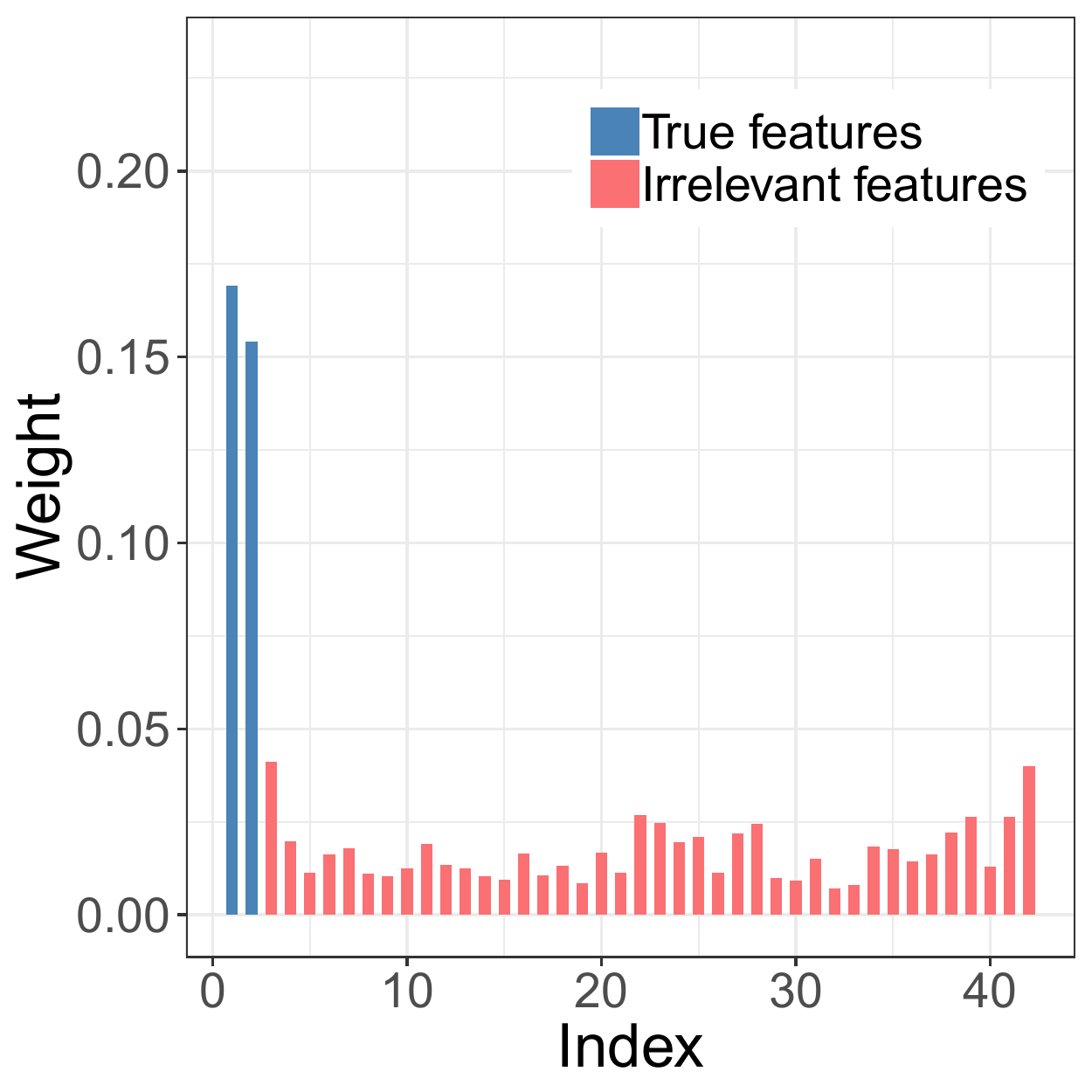}
\label{Fig:05b}
\end{minipage}}
\hspace{4mm}
\subfigure[Select-ProtoNet]{
\begin{minipage}[t]{0.28\linewidth}
\centering
\includegraphics[width=1.4in]{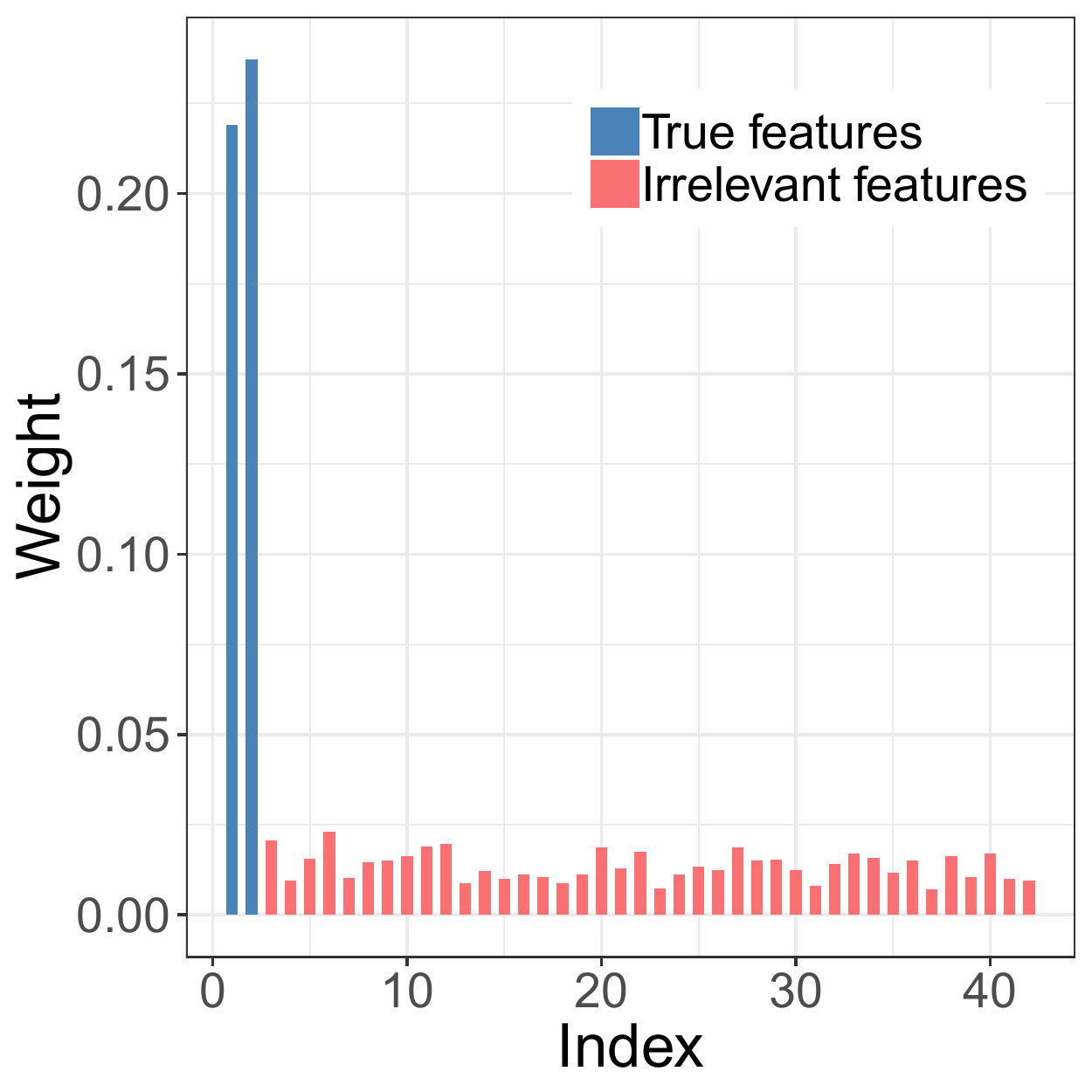}
\label{Fig:05c}
\end{minipage}}\vspace{-2mm}
\caption{Plots of feature weights learned by the competing methods on unbiased simulated data. The red bar represents the true features and the blue bar represents the irrelevant features.}
\label{Fig:05}
\end{figure}

\textbf{Results.}
Table~\ref{Tab:01} shows the accuracy of Select-ProtoNet and its backbones on simulated datasets under different experiment settings with $30$ random runs. It can be seen that the appended two modules both make contributions to improve the performance, and the proposed method achieves the best results. With the increase of noise rate, the advantages of Select-ProtoNet over ProtoNet become more obvious. As shown in Fig.~\ref{Fig:03}, the curves of training loss and accuracy comparison between ProtoNet and Select-ProtoNet. We can observe that ProtoNet takes almost twice the time of our model to achieve the best classification accuracy. To illustrate the effect of two appended modules of our model, we plot the weight distribution of clean and noisy training support samples in Fig.~\ref{Fig:07}. It can be seen that almost all lager weights belong to clean samples, and the weight value of noisy samples is less than that of clean samples, which implies that MW-NetV2 can distinguish clean and noisy samples. Fig.~\ref{Fig:05c} shows the feature weights learned by Select-ProtoNet. It can be observed that the weights of important features are higher than those of irrelevant features, and almost all the weights of irrelevant features are less than a certain threshold value, indicating that the feature selection layer can select important features.

We also compare Select-ProtoNet with state of the art conventional methods. Fig.~\ref{Fig:04} shows the test accuracy of all competing methods on the few selected training samples (only 1\% of datasets). It can be seen that our model is significantly superior to all other competing methods. There are big performance gaps between our model and conventional methods (i.e., Random Forest, NeuralNet, SVM, and Logistic Regression) when training data is small. Particularly, NeuralNet, is the worst performer because the training pool is quite small that the power of deep learning can only be manifested when a large amount of data is available. In addition, we plot the weights of features learned by NeuralNet and AffinityNet, are shown in Fig~\ref{Fig:05}. We can observe that feature selection performance of our model outperforms competing methods. We further increase the feature dimension to evaluate the classification accuracy of all comparison methods. The results are shown in Table~\ref{Tab:02} with $30$ random runs. As can be seen, Select-ProtoNet can improve the accuracy compared with ProtoNet, and outperforms all other conventional comparison methods. With the increase of the feature dimension, the performance gaps between the proposed method and all competing methods increase gradually.

\begin{table}[!pbt]
\caption{Few-shot classification accuracy (\%) comparison on simulated data with varying experiment settings. The best results are highlighted in \textbf{bold}.}
\label{Tab:01}
\centering
\footnotesize
\renewcommand\arraystretch{1.05}
\begin{tabular}{c|c|cccc}
\hline
\textbf{Noise   rate} & \textbf{Model}   & \textbf{5 way 5 shot} & \textbf{5 way 10 shot} & \textbf{10 way 5 shot} & \textbf{10 way 10 shot} \\ \hline
\multirow{4}{*}{0\%}  & ProtoNet         & 91.93 $\pm$ 3.08          & 92.30 $\pm$ 2.61           & 93.01 $\pm$ 1.81           & 93.40 $\pm$ 1.30            \\
                      & SelectS-ProtoNet & 92.74 $\pm$ 1.42          & 93.29 $\pm$ 1.39           & 93.70 $\pm$ 1.21           & 94.30 $\pm$ 1.40            \\
                      & SelectF-ProtoNet & 96.55 $\pm$ 1.75          & 96.78 $\pm$ 1.02           & 97.42 $\pm$ 0.48           & 97.21 $\pm$ 0.51            \\
                      & Select-ProtoNet  & \textbf{96.69 $\pm$ 1.08} & \textbf{96.94 $\pm$ 0.98}  & \textbf{97.44 $\pm$ 0.41}  & \textbf{97.62 $\pm$ 0.47}   \\ \hline
\multirow{4}{*}{10\%} & ProtoNet         & 83.23 $\pm$ 4.90          & 82.25 $\pm$ 4.10           & 85.27 $\pm$ 3.41           & 86.38 $\pm$ 4.00            \\
                      & SelectS-ProtoNet & 84.30 $\pm$ 4.44          & 84.49 $\pm$ 3.73           & 88.26 $\pm$ 2.81           & 89.89 $\pm$ 2.59            \\
                      & SelectF-ProtoNet & 91.17 $\pm$ 3.59          & 92.29 $\pm$ 2.05           & 92.62 $\pm$ 2.24           & 92.96 $\pm$ 2.00            \\
                      & Select-ProtoNet  & \textbf{93.56 $\pm$ 2.78} & \textbf{94.40 $\pm$ 1.34}  & \textbf{94.67 $\pm$ 1.55}  & \textbf{94.31 $\pm$ 1.73}   \\ \hline
\multirow{4}{*}{30\%} & ProtoNet         & 72.33 $\pm$ 7.42          & 73.98  $\pm$ 6.59          & 74.02 $\pm$ 6.43           & 74.52 $\pm$ 6.13            \\
                      & SelectS-ProtoNet & 75.74 $\pm$ 6.21          & 76.34 $\pm$ 4.92           & 77.64 $\pm$ 7.53           & 77.81 $\pm$ 6.00            \\
                      & SelectF-ProtoNet & 86.56 $\pm$ 4.48          & 87.07 $\pm$ 5.49           & 87.72 $\pm$ 4.53           & 86.29 $\pm$ 6.22            \\
                      & Select-ProtoNet  & \textbf{88.95 $\pm$ 3.74} & \textbf{89.66 $\pm$ 4.02}  & \textbf{89.87 $\pm$ 3.85}  & \textbf{89.60 $\pm$ 4.38}   \\ \hline
\multirow{4}{*}{50\%} & ProtoNet         & 64.35 $\pm$ 7.99          & 65.95 $\pm$ 8.01           & 66.93 $\pm$ 7.60           & 66.84 $\pm$ 6.86            \\
                      & SelectS-ProtoNet & 68.69 $\pm$ 6.42          & 68.79 $\pm$ 7.99           & 70.22 $\pm$ 7.75           & 70.90 $\pm$ 6.30            \\
                      & SelectF-ProtoNet & 79.04 $\pm$ 9.33          & 80.85 $\pm$ 7.44           & 81.30 $\pm$ 7.08           & 81.47 $\pm$ 8.95            \\
                      & Select-ProtoNet  & \textbf{83.42 $\pm$ 6.06} & \textbf{83.21 $\pm$ 5.59}  & \textbf{84.49 $\pm$ 5.33}  & \textbf{84.43 $\pm$ 5.54}   \\ \hline
\end{tabular}
\end{table}

 \vspace{-4mm}

\vspace{-2mm}

\begin{table}[!pbt]
\caption{Test accuracy (\%) comparison on simulated data with varying feature dimension settings. The best results are highlighted in \textbf{bold}.}
\label{Tab:02}
\centering
\footnotesize
\renewcommand\arraystretch{1}
\begin{tabular}{c|cccc}
\hline
Methods               & 100                   & 500                   & 1000                  & 2000                  \\ \hline
SVM                   & 40.79 $\pm$ 1.47          & 30.84 $\pm$ 0.82          & 29.34 $\pm$ 1.08          & 27.95 $\pm$ 0.73          \\
Naive Bayes           & 76.91 $\pm$ 5.03          & 54.19 $\pm$ 5.34          & 48.17 $\pm$ 3.07          & 40.47 $\pm$ 3.10          \\
Random Forest         & 52.23 $\pm$ 9.47          & 26.43 $\pm$ 2.48          & 26.21 $\pm$ 2.71          & 25.73 $\pm$ 1.63          \\
Logistic   Regression & 41.59 $\pm$ 1.86          & 30.83 $\pm$ 0.92          & 29.36 $\pm$ 1.07          & 27.97 $\pm$ 0.56          \\
Logistic\_lasso       & 72.96 $\pm$ 4.72          & 59.24 $\pm$ 4.78          & 49.65 $\pm$ 5.94          & 42.31 $\pm$ 5.60          \\
NeuralNet             & 32.42 $\pm$ 1.50          & 27.61 $\pm$ 0.75          & 26.29 $\pm$ 0.85          & 25.92 $\pm$ 0.66          \\
AffinityNet           & 66.06 $\pm$ 25.13         & 39.17 $\pm$ 13.30         & 27.55 $\pm$ 6.98          & 34.18 $\pm$ 10.66         \\ \hline
ProtoNet              & 80.48 $\pm$ 4.68          & 71.90 $\pm$ 2.64          & 54.09 $\pm$ 10.95         & 48.66 $\pm$ 6.29          \\
Select-ProtoNet       & \textbf{96.08 $\pm$ 0.35} & \textbf{93.13 $\pm$ 2.44} & \textbf{85.40 $\pm$ 5.12} & \textbf{81.49 $\pm$ 6.10} \\ \hline
\end{tabular}
\end{table}

\subsection{Experiments on \emph{mini}TCGA Meta-Dataset}
\textbf{Real Datasets.}
TCGA Meta-Dataset \cite{samiei2019tcga} is a publicly available benchmark dataset in the field of gene expression analysis, containing 174 clinical tasks derived from The Cancer Genome Atlas (TCGA) \cite{liu2018integrated}, which can be used in a multi-task learning framework. However, it cannot be directly used as a few-shot learning benchmark dataset because some tasks have extremely imbalance classes. Thus, after carefully selecting samples, we devise a new dataset, \emph{mini}-TCGA Meta-Dataset, consisting of 68 TCGA benchmark clinical tasks, where each task has two classes and each class has at least 60 samples. More details about datasets can be found in Appendix D.

\textbf{Baselines.}
We compare against the following state-of-art methods: ProtoNet, the majority class prediction (Majority), Logistic Regression, and Neural Network. The implementations of three conventional methods are the same as the work in \cite{samiei2019tcga}. For details on implementations, training and test procedures, see Appendix E.

\textbf{Results.}
The classification accuracy of \emph{mini}TCGA Meta-Dataset with varying noise rate settings are reported in Table~\ref{Tab:03}. As can be seen, our model improves on its backbone method with a large margin for all noise rate settings, and outperforms the three conventional supervised methods.

\begin{wrapfigure}{r}{0.3\textwidth}
	\begin{center}
	\includegraphics[width=0.3\textwidth]{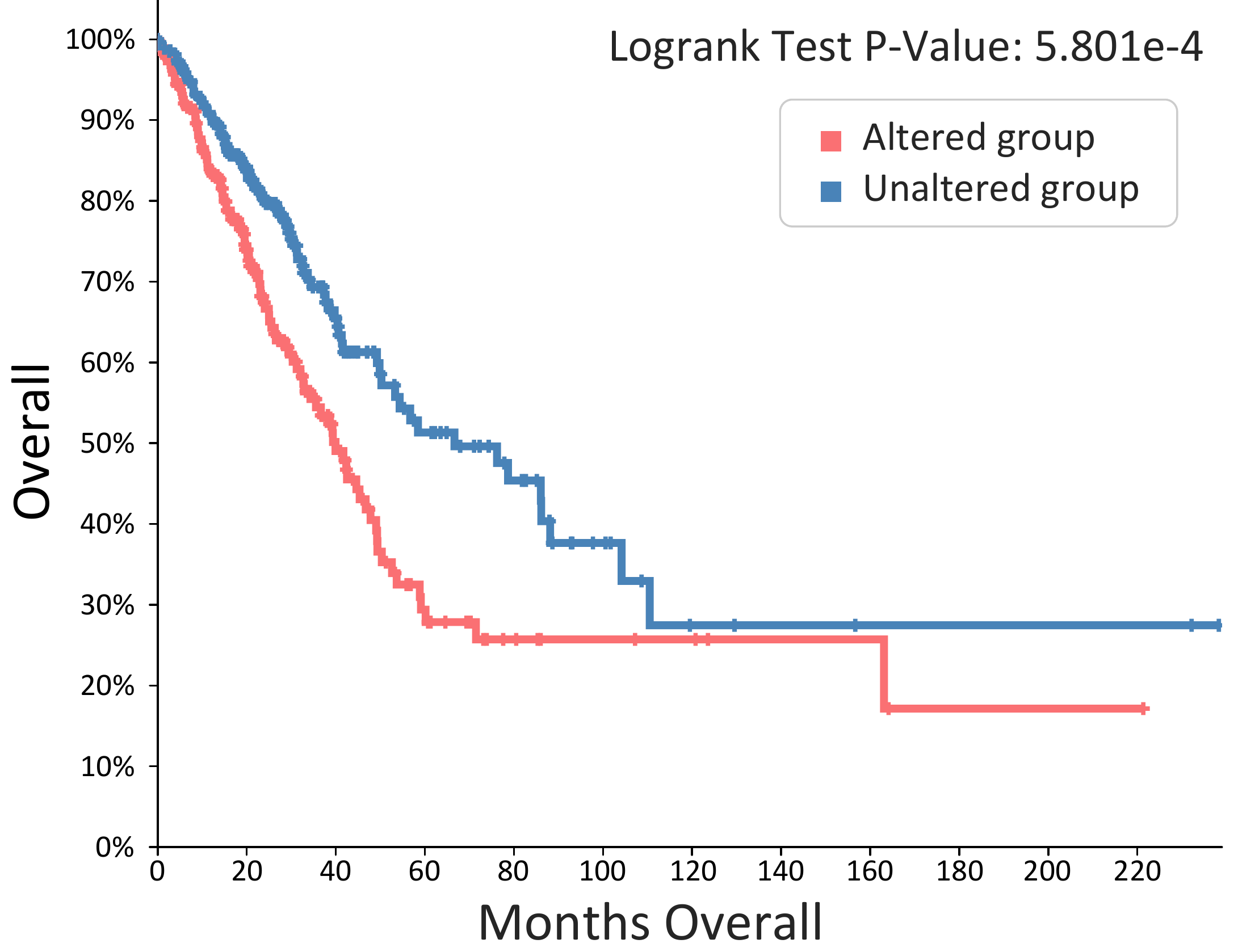}
	\end{center}	
	\vspace{-4mm}
	\caption{Survival curves of the 20 top-ranked genes selected by Select-ProtoNet.}
	\label{Fig:06}
	\vspace{-4mm}
\end{wrapfigure}
Table~\ref{Tab:04} shows the lung cancer subtype prediction accuracy of all competing methods on the TCGA Meta-Dataset with task id (`Expression\_Subtype', `LUNG'). Especially, ProtoNet and Select-ProtoNet consider the interrelationship of clinical tasks, leveraging samples from all clinical tasks of \emph{mini}TCGA Meta-Dataset to build shared experience or knowledge and transfer it to help predict the lung cancer subtype. The conventional methods, however, only consider the lung cancer subtype task. As shown in Table~\ref{Tab:04}, we can observe that our model achieves the best results, with an accuracy gain of more than 20\% compared with its backbone, and over 29\% against the best result of the conventional methods. It implies that our model significantly improves the prediction performance of disease subtypes by incorporating various interrelated clinical tasks. Fig.~\ref{Fig:06} shows the Kaplan-Meier (KM) survival curves of the 20 top-ranked significant genes selected by Select-ProtoNet on the lung cancer subtype task of TCGA Meta-Dataset. By using bBioPortal \cite{cerami2012cbio, gao2013integrative}, the survival analysis of significant genes is done based on the Pan-Cancer Atlas dataset \cite{hoadley2018cell}. As shown in Fig.~\ref{Fig:06}, the two curves without intersect (logrank test $p$-value $=5.801e-4$). The patients without alterations in the selected genes (blue line) have long survival times.

\begin{table}[!pbt]
\caption{Test accuracy (\%) comparison on \emph{mini}TCGA Meta-Dataset with varying noise rate settings. The mean accuracy ($\pm$ std) over 30 repetitions are reported. The best results are highlighted in \textbf{bold}.} \vspace{-2mm}
\label{Tab:03}
\centering
\footnotesize
\renewcommand\arraystretch{1.05}
\resizebox{\textwidth}{!}{
\begin{tabular}{c|ccc|cc}
\hline
\multirow{2}{*}{Noise   Rate} & \multicolumn{3}{c|}{Conventional   supervised methods} & \multicolumn{2}{c}{Select-ProtoNet and   its backbone} \\ \cline{2-6}
                              & Majority      & Logistic Regression  & Neural Network & ProtoNet               & Select-ProtoNet               \\ \hline
0\%                           & 64.10 $\pm$ 8.83  & 67.96 $\pm$ 12.80        & 68.30 $\pm$ 11.63  & 72.01 $\pm$ 19.91          & \textbf{84.71 $\pm$ 4.49}         \\
10\%                          & 63.63 $\pm$ 9.38  & 65.08 $\pm$ 10.49        & 64.85 $\pm$ 9.98   & 66.69 $\pm$ 19.54          & \textbf{81.33 $\pm$ 2.32}         \\
30\%                          & 61.98 $\pm$ 9.94  & 60.89 $\pm$ 8.00         & 61.50 $\pm$ 8.89   & 64.18 $\pm$ 19.48          & \textbf{79.89 $\pm$ 2.40}         \\
50\%                          & 59.73 $\pm$ 9.56  & 57.06 $\pm$ 5.72         & 58.73 $\pm$ 7.59   & 60.72 $\pm$ 17.31          & \textbf{78.15 $\pm$ 2.30}         \\ \hline
\end{tabular}}
\end{table}

\begin{table}[!pbt]
\centering
\caption{Accuracy (\%) comparison on the lung cancer subtype task of TCGA Meta-Dataset.}\vspace{-2mm}
\label{Tab:04}
\renewcommand\arraystretch{1.05}
\resizebox{\textwidth}{!}{
\begin{tabular}{c|ccccc}
\hline
Method   & Majority     & Logistic Regression & Neural Network & ProtoNet      & Select-ProtoNet       \\ \hline
Accuracy & 38.00 $\pm$ 0.00 & 60.64 $\pm$ 8.23        & 68.20 $\pm$ 1.60   & 77.78 $\pm$ 27.22 & \textbf{97.78 $\pm$ 3.44} \\ \hline
\end{tabular}}
\end{table}

\section{Conclusion}
In this paper, we propose a novel meta learning method for few-shot disease subtype prediction problem. The proposed method, carefully designs to append two modules to the vanilla Prototypical Network, making it able to address the high dimensionality and high noise issues in gene expression data compared with image data. Meanwhile, it possesses the ability of Prototypical Network to
extract the shared experience or knowledge from interrelated clinical tasks, capable of learning from a limited number of training examples and generalize well.
Synthetic and a new released benchmark dataset for disease subtype prediction, \emph{mini}-TCGA Meta-Dataset experiments
substantiate the superiority of the proposed method for predicting the subtypes of disease and identifying potential disease-related genes.
In future work, we need to further consider the interaction between genes and integrate gene regulatory networks into meta-learning techniques and further promote the application of meta-learning techniques in the field of few-shot biological genomics, thus bringing us closer to the clinic of the future.

\small
\bibliographystyle{plain}
\bibliography{Reference}

\begin{thebibliography}{10}

\bibitem{bengio1992optimization}
Samy Bengio, Yoshua Bengio, Jocelyn Cloutier, and Jan Gecsei.
\newblock On the optimization of a synaptic learning rule.
\newblock In {\em Preprints Conf. Optimality in Artificial and Biological
  Neural Networks}, volume~2. Univ. of Texas, 1992.

\bibitem{cerami2012cbio}
Ethan Cerami, Jianjiong Gao, Ugur Dogrusoz, Benjamin~E Gross, Selcuk~Onur
  Sumer, B{\"u}lent~Arman Aksoy, Anders Jacobsen, Caitlin~J Byrne, Michael~L
  Heuer, Erik Larsson, et~al.
\newblock The cbio cancer genomics portal: an open platform for exploring
  multidimensional cancer genomics data, 2012.

\bibitem{de2003framework}
Fernando De~La~Torre and Michael~J Black.
\newblock A framework for robust subspace learning.
\newblock {\em International Journal of Computer Vision}, 54(1-3):117--142,
  2003.

\bibitem{dehghani2017fidelity}
Mostafa Dehghani, Arash Mehrjou, Stephan Gouws, Jaap Kamps, and Bernhard
  Sch{\"o}lkopf.
\newblock Fidelity-weighted learning.
\newblock {\em arXiv preprint arXiv:1711.02799}, 2017.

\bibitem{finn2017model}
Chelsea Finn, Pieter Abbeel, and Sergey Levine.
\newblock Model-agnostic meta-learning for fast adaptation of deep networks.
\newblock In {\em Proceedings of the 34th International Conference on Machine
  Learning-Volume 70}, pages 1126--1135. JMLR. org, 2017.

\bibitem{finn2018probabilistic}
Chelsea Finn, Kelvin Xu, and Sergey Levine.
\newblock Probabilistic model-agnostic meta-learning.
\newblock In {\em Advances in Neural Information Processing Systems}, pages
  9516--9527, 2018.

\bibitem{gao2013integrative}
Jianjiong Gao, B{\"u}lent~Arman Aksoy, Ugur Dogrusoz, Gideon Dresdner, Benjamin
  Gross, S~Onur Sumer, Yichao Sun, Anders Jacobsen, Rileen Sinha, Erik Larsson,
  et~al.
\newblock Integrative analysis of complex cancer genomics and clinical profiles
  using the cbioportal.
\newblock {\em Sci. Signal.}, 6(269):pl1--pl1, 2013.

\bibitem{hoadley2018cell}
Katherine~A Hoadley, Christina Yau, Toshinori Hinoue, Denise~M Wolf,
  Alexander~J Lazar, Esther Drill, Ronglai Shen, Alison~M Taylor, Andrew~D
  Cherniack, V{\'e}steinn Thorsson, et~al.
\newblock Cell-of-origin patterns dominate the molecular classification of
  10,000 tumors from 33 types of cancer.
\newblock {\em Cell}, 173(2):291--304, 2018.

\bibitem{hospedales2020meta}
Timothy Hospedales, Antreas Antoniou, Paul Micaelli, and Amos Storkey.
\newblock Meta-learning in neural networks: A survey.
\newblock {\em arXiv preprint arXiv:2004.05439}, 2020.

\bibitem{hughey2015robust}
Jacob~J Hughey and Atul~J Butte.
\newblock Robust meta-analysis of gene expression using the elastic net.
\newblock {\em Nucleic acids research}, 43(12):e79--e79, 2015.

\bibitem{jiang2014easy}
Lu~Jiang, Deyu Meng, Teruko Mitamura, and Alexander~G Hauptmann.
\newblock Easy samples first: Self-paced reranking for zero-example multimedia
  search.
\newblock In {\em Proceedings of the 22nd ACM international conference on
  Multimedia}, pages 547--556, 2014.

\bibitem{kumar2010self}
M~Pawan Kumar, Benjamin Packer, and Daphne Koller.
\newblock Self-paced learning for latent variable models.
\newblock In {\em Advances in Neural Information Processing Systems}, pages
  1189--1197, 2010.

\bibitem{lazar2013batch}
Cosmin Lazar, Stijn Meganck, Jonatan Taminau, David Steenhoff, Alain Coletta,
  Colin Molter, David~Y Weiss-Sol{\'\i}s, Robin Duque, Hugues Bersini, and Ann
  Now{\'e}.
\newblock Batch effect removal methods for microarray gene expression data
  integration: a survey.
\newblock {\em Briefings in bioinformatics}, 14(4):469--490, 2013.

\bibitem{lazar2012survey}
Cosmin Lazar, Jonatan Taminau, Stijn Meganck, David Steenhoff, Alain Coletta,
  Colin Molter, Virginie de~Schaetzen, Robin Duque, Hugues Bersini, and Ann
  Nowe.
\newblock A survey on filter techniques for feature selection in gene
  expression microarray analysis.
\newblock {\em IEEE/ACM Transactions on Computational Biology and
  Bioinformatics}, 9(4):1106--1119, 2012.

\bibitem{lee2018gradient}
Yoonho Lee and Seungjin Choi.
\newblock Gradient-based meta-learning with learned layerwise metric and
  subspace.
\newblock {\em arXiv preprint arXiv:1801.05558}, 2018.

\bibitem{leek2010tackling}
Jeffrey~T Leek, Robert~B Scharpf, H{\'e}ctor~Corrada Bravo, David Simcha,
  Benjamin Langmead, W~Evan Johnson, Donald Geman, Keith Baggerly, and Rafael~A
  Irizarry.
\newblock Tackling the widespread and critical impact of batch effects in
  high-throughput data.
\newblock {\em Nature Reviews Genetics}, 11(10):733--739, 2010.

\bibitem{li2014meta}
Quefeng Li, Sijian Wang, Chiang-Ching Huang, Menggang Yu, and Jun Shao.
\newblock Meta-analysis based variable selection for gene expression data.
\newblock {\em Biometrics}, 70(4):872--880, 2014.

\bibitem{liang2013sparse}
Yong Liang, Cheng Liu, Xin-Ze Luan, Kwong-Sak Leung, Tak-Ming Chan, Zong-Ben
  Xu, and Hai Zhang.
\newblock Sparse logistic regression with a l 1/2 penalty for gene selection in
  cancer classification.
\newblock {\em BMC bioinformatics}, 14(1):198, 2013.

\bibitem{liu2018integrated}
Jianfang Liu, Tara Lichtenberg, Katherine~A Hoadley, Laila~M Poisson,
  Alexander~J Lazar, Andrew~D Cherniack, Albert~J Kovatich, Christopher~C Benz,
  Douglas~A Levine, Adrian~V Lee, et~al.
\newblock An integrated tcga pan-cancer clinical data resource to drive
  high-quality survival outcome analytics.
\newblock {\em Cell}, 173(2):400--416, 2018.

\bibitem{ma2019affinitynet}
Tianle Ma and Aidong Zhang.
\newblock Affinitynet: semi-supervised few-shot learning for disease type
  prediction.
\newblock In {\em Proceedings of the AAAI Conference on Artificial
  Intelligence}, volume~33, pages 1069--1076, 2019.

\bibitem{mandanas2020subspace}
Fotios~D Mandanas and Constantine~L Kotropoulos.
\newblock Subspace learning and feature selection via orthogonal mapping.
\newblock {\em IEEE Transactions on Signal Processing}, 2020.

\bibitem{raser2005noise}
Jonathan~M Raser and Erin~K O'Shea.
\newblock Noise in gene expression: origins, consequences, and control.
\newblock {\em Science}, 309(5743):2010--2013, 2005.

\bibitem{ravi2016optimization}
Sachin Ravi and Hugo Larochelle.
\newblock Optimization as a model for few-shot learning.
\newblock 2016.

\bibitem{ren2018meta}
Mengye Ren, Eleni Triantafillou, Sachin Ravi, Jake Snell, Kevin Swersky,
  Joshua~B Tenenbaum, Hugo Larochelle, and Richard~S Zemel.
\newblock Meta-learning for semi-supervised few-shot classification.
\newblock {\em arXiv preprint arXiv:1803.00676}, 2018.

\bibitem{ren2018learning}
Mengye Ren, Wenyuan Zeng, Bin Yang, and Raquel Urtasun.
\newblock Learning to reweight examples for robust deep learning.
\newblock {\em arXiv preprint arXiv:1803.09050}, 2018.

\bibitem{rhodes2005integrative}
Daniel~R Rhodes and Arul~M Chinnaiyan.
\newblock Integrative analysis of the cancer transcriptome.
\newblock {\em Nature genetics}, 37(6s):S31, 2005.

\bibitem{rhodes2004large}
Daniel~R Rhodes, Jianjun Yu, K~Shanker, Nandan Deshpande, Radhika Varambally,
  Debashis Ghosh, Terrence Barrette, Akhilesh Pandey, and Arul~M Chinnaiyan.
\newblock Large-scale meta-analysis of cancer microarray data identifies common
  transcriptional profiles of neoplastic transformation and progression.
\newblock {\em Proceedings of the National Academy of Sciences},
  101(25):9309--9314, 2004.

\bibitem{samiei2019tcga}
Mandana Samiei, Tobias W{\"u}rfl, Tristan Deleu, Martin Weiss, Francis Dutil,
  Thomas Fevens, Genevi{\`e}ve Boucher, Sebastien Lemieux, and Joseph~Paul
  Cohen.
\newblock The tcga meta-dataset clinical benchmark.
\newblock {\em arXiv preprint arXiv:1910.08636}, 2019.

\bibitem{saria2015subtyping}
Suchi Saria and Anna Goldenberg.
\newblock Subtyping: What it is and its role in precision medicine.
\newblock {\em IEEE Intelligent Systems}, 30(4):70--75, 2015.

\bibitem{schmidhuber1987evolutionary}
J{\"u}rgen Schmidhuber.
\newblock {\em Evolutionary principles in self-referential learning, or on
  learning how to learn: the meta-meta-... hook}.
\newblock PhD thesis, Technische Universit{\"a}t M{\"u}nchen, 1987.

\bibitem{schmiedel2019empirical}
J{\"o}rn~M Schmiedel, Lucas~B Carey, and Ben Lehner.
\newblock Empirical mean-noise fitness landscapes reveal the fitness impact of
  gene expression noise.
\newblock {\em Nature communications}, 10(1):1--12, 2019.

\bibitem{shabalin2008merging}
Andrey~A Shabalin, H{\aa}kon Tjelmeland, Cheng Fan, Charles~M Perou, and
  Andrew~B Nobel.
\newblock Merging two gene-expression studies via cross-platform normalization.
\newblock {\em Bioinformatics}, 24(9):1154--1160, 2008.

\bibitem{shu2019meta}
Jun Shu, Qi~Xie, Lixuan Yi, Qian Zhao, Sanping Zhou, Zongben Xu, and Deyu Meng.
\newblock Meta-weight-net: Learning an explicit mapping for sample weighting.
\newblock {\em arXiv preprint arXiv:1902.07379}, 2019.

\bibitem{shu2018small}
Jun Shu, Zongben Xu, and Deyu Meng.
\newblock Small sample learning in big data era.
\newblock {\em arXiv preprint arXiv:1808.04572}, 2018.

\bibitem{snell2017prototypical}
Jake Snell, Kevin Swersky, and Richard Zemel.
\newblock Prototypical networks for few-shot learning.
\newblock In {\em Advances in Neural Information Processing Systems}, pages
  4077--4087, 2017.

\bibitem{sohn2017clinical}
Bo~Hwa Sohn, Jun-Eul Hwang, Hee-Jin Jang, Hyun-Sung Lee, Sang~Cheul Oh, Jae-Jun
  Shim, Keun-Wook Lee, Eui~Hyun Kim, Sun~Young Yim, Sang~Ho Lee, et~al.
\newblock Clinical significance of four molecular subtypes of gastric cancer
  identified by the cancer genome atlas project.
\newblock {\em Clinical Cancer Research}, 23(15):4441--4449, 2017.

\bibitem{thrun1998lifelong}
Sebastian Thrun.
\newblock Lifelong learning algorithms.
\newblock In {\em Learning to learn}, pages 181--209. Springer, 1998.

\bibitem{tibshirani1996regression}
Robert Tibshirani.
\newblock Regression shrinkage and selection via the lasso.
\newblock {\em Journal of the Royal Statistical Society: Series B
  (Methodological)}, 58(1):267--288, 1996.

\bibitem{vinyals2016matching}
Oriol Vinyals, Charles Blundell, Timothy Lillicrap, Daan Wierstra, et~al.
\newblock Matching networks for one shot learning.
\newblock In {\em Advances in neural information processing systems}, pages
  3630--3638, 2016.

\bibitem{wang2014similarity}
Bo~Wang, Aziz~M Mezlini, Feyyaz Demir, Marc Fiume, Zhuowen Tu, Michael Brudno,
  Benjamin Haibe-Kains, and Anna Goldenberg.
\newblock Similarity network fusion for aggregating data types on a genomic
  scale.
\newblock {\em Nature methods}, 11(3):333, 2014.

\bibitem{wang2018low}
Yu-Xiong Wang, Ross Girshick, Martial Hebert, and Bharath Hariharan.
\newblock Low-shot learning from imaginary data.
\newblock In {\em Proceedings of the IEEE conference on computer vision and
  pattern recognition}, pages 7278--7286, 2018.

\bibitem{wirapati2008meta}
Pratyaksha Wirapati, Christos Sotiriou, Susanne Kunkel, Pierre Farmer, Sylvain
  Pradervand, Benjamin Haibe-Kains, Christine Desmedt, Michail Ignatiadis,
  Thierry Sengstag, Fr{\'e}d{\'e}ric Sch{\"u}tz, et~al.
\newblock Meta-analysis of gene expression profiles in breast cancer: toward a
  unified understanding of breast cancer subtyping and prognosis signatures.
\newblock {\em Breast Cancer Research}, 10(4):R65, 2008.

\bibitem{xing2019adaptive}
Chen Xing, Negar Rostamzadeh, Boris Oreshkin, and Pedro~OO Pinheiro.
\newblock Adaptive cross-modal few-shot learning.
\newblock In {\em Advances in Neural Information Processing Systems}, pages
  4848--4858, 2019.

\bibitem{xu20101}
Zongben Xu, Hai Zhang, Yao Wang, XiangYu Chang, and Yong Liang.
\newblock L 1/2 regularization.
\newblock {\em Science China Information Sciences}, 53(6):1159--1169, 2010.

\bibitem{xue2015survey}
Bing Xue, Mengjie Zhang, Will~N Browne, and Xin Yao.
\newblock A survey on evolutionary computation approaches to feature selection.
\newblock {\em IEEE Transactions on Evolutionary Computation}, 20(4):606--626,
  2015.

\bibitem{yang2019multi}
Zi-Yi Yang, Xiao-Ying Liu, Jun Shu, Hui Zhang, Yan-Qiong Ren, Zong-Ben Xu, and
  Yong Liang.
\newblock Multi-view based integrative analysis of gene expression data for
  identifying biomarkers.
\newblock {\em Scientific reports}, 9(1):1--15, 2019.

\bibitem{yang2019mspl}
Zi-Yi Yang, Liang-Yong Xia, Hui Zhang, and Yong Liang.
\newblock Mspl: Multimodal self-paced learning for multi-omics feature
  selection and data integration.
\newblock {\em IEEE Access}, 7:170513--170524, 2019.

\bibitem{zhang2018generalized}
Zhilu Zhang and Mert Sabuncu.
\newblock Generalized cross entropy loss for training deep neural networks with
  noisy labels.
\newblock In {\em Advances in neural information processing systems}, pages
  8778--8788, 2018.

\bibitem{zou2005regularization}
Hui Zou and Trevor Hastie.
\newblock Regularization and variable selection via the elastic net.
\newblock {\em Journal of the royal statistical society: series B (statistical
  methodology)}, 67(2):301--320, 2005.

\end{thebibliography}

\end{document}